\documentclass[11pt]{article}

\usepackage[preprint]{acl}

\usepackage{times}
\usepackage{latexsym}

\usepackage[T1]{fontenc}

\usepackage[utf8]{inputenc}

\usepackage{microtype}

\usepackage{inconsolata}

\usepackage{graphicx}

\usepackage{xcolor}
\usepackage{url}
\usepackage{algorithm}
\usepackage{algpseudocode}
\usepackage{tabularx} 
\usepackage{amsmath}
\usepackage{xspace}
\usepackage{booktabs}
\usepackage{amssymb}
\usepackage{makecell}
\usepackage{multirow}
\usepackage{multicol}
\usepackage{arydshln}
\usepackage{subcaption}
\usepackage{listings}
\usepackage[most]{tcolorbox}
\newcommand{\method}{SkillGrad\xspace}
\newcommand{\scorestd}[1]{\mbox{\scriptsize $\pm$ #1}}

\title{SkillGrad: Optimizing Agent Skills Like Gradient Descent}

\author{
  \textbf{Hanyu Wang},
  \textbf{Yifan Lan},
  \textbf{Bochuan Cao},\\
  \textbf{Lu Lin},
  \textbf{Jinghui Chen}
\\
\\
  College of Information Sciences and Technology \\
  The Pennsylvania State University\\
  University Park, PA, USA
\\
  \small{
    \textbf{Correspondence to:}
    \href{mailto:hbw5365@psu.edu}{hbw5365@psu.edu};
    \href{mailto:jzc5917@psu.edu}{jzc5917@psu.edu}
  }
}

\begin{document}
\maketitle
\begin{abstract}
Agent skills provide a lightweight way to adapt LLM agents to specialized domains by storing reusable procedural knowledge in structured files. However, whether downloaded from third parties or self-generated, these skills are often unreliable, incomplete, or outdated. Existing skill-evolution methods often address these deficiencies through heuristic reflections without an explicit optimization formulation. In this paper, we propose \textbf{\method}, a gradient-descent-inspired framework for optimizing agent skills. \method treats the skill package as a structured parameter to optimize in a gradient descent fashion: task executions provide trajectory-level loss evidence, automatic diagnoses then provide text-based gradients that indicate the correction directions. To stabilize optimization across iterations, a momentum agent accumulates recurring diagnostic patterns into a persistent memory overlay. Finally, an LLM-based patcher executes the parameter update by applying layer-aware edits to the skill package. Evaluated on SpreadsheetBench Verified and WikiTableQuestions, \method consistently outperforms training-based skill evolution baselines across two backbone LLMs, improving over the strongest training-based baseline by $6.7$ percentage points on average. Ablations further show that momentum and contrastive diagnosis both contribute to the final skill quality. Code will be updated at \url{https://github.com/wwwhy725/SkillGrad}
\end{abstract}

\section{Introduction}
\label{sec:intro}

Large Language Model agents \citep{yao2022react, wang2023voyager} have evolved rapidly, achieving impressive proficiency in long-horizon decision-making tasks such as reasoning \citep{chang2026memcollab, xi2025agentgym, lan2026illusion}, planning \citep{erdogan2025plan, wang2026preflect}, and web navigation \citep{he2024webvoyager, deng2023mind2web, wu2026webdancer}. 
However, many practical agent applications require more than general problem-solving ability. In specialized, procedure-heavy domains, such as spreadsheet manipulation \citep{chen2024sheetagent}, document editing \citep{li2025aid}, and codebase maintenance \citep{li2026environment}, agents must repeatedly follow domain-specific workflows, use specialized tools correctly, and handle recurring edge cases. Adapting agents to various domains through fine-tuning \citep{liu2024moe, chang2026enhancing}, retrieval pipelines \citep{zhao2025medrag}, or repeated web searches \citep{shao2024assisting} can be costly or cumbersome, especially when the needed knowledge is procedural rather than purely factual. To bridge this gap, \textit{Agent Skills} offer a lightweight alternative. They are persistent file packages that an agent can load progressively when solving tasks. Unlike a flat prompt, a skill is a structured artifact. Its metadata determines when it is activated, its \texttt{SKILL.md} body is always loaded after activation, and additional resources are consulted only when relevant.

However, the usefulness of this adaptation depends critically on skill quality. SkillsBench \citep{li2026skillsbench} shows that automatically generated skills can remain well below expert-written ones, and in some cases even degrade agent performance relative to using no skill. This problem is broader than automatic skill generation, because any fixed skill package can omit task-specific edge cases, become misaligned with the target task distribution, or encode brittle assumptions about tools and workflows. Such problems motivate a natural question: \textit{can we treat a skill as an optimizable artifact and systematically improve it after initialization?}

To answer this question, we introduce \textbf{\method}, a gradient-descent-inspired framework for optimizing agent skills. The correspondence is conceptual rather than numeric. Agent skills are discrete text artifacts, so there is no literal derivative. Instead, the analogy provides a principled design lens, summarized in Table~\ref{tab:analogy}. The parameter is the structured skill package $S_t$. At each iteration, the current skill is executed on a mini-batch of tasks, producing outcomes $r_{t,i}$ and trajectories $\tau_{t,i}$ as loss evidence. A diagnoser converts this evidence into textual update signals $d_{t,i}$, analogous to per-example gradients. Failed trajectories expose corrective changes, while contrastive successful trajectories, where the initial skill failed but the current skill succeeds, identify behaviors worth preserving. A momentum agent accumulates recurring patterns into a persistent memory $M_t$ and a current overlay $O_t$, and a patcher applies a layer-aware edit to obtain the next skill package $S_{t+1}$.

We evaluate \method on SpreadsheetBench Verified \citep{ma2024spreadsheetbench} and WikiTableQuestions \citep{pasupat2015compositional} using two backbone LLMs and two sources of initial \texttt{xlsx} skills, one generated by an LLM and one downloaded from a third party. Both cases share the same goal of optimizing the given skill package under a fixed configuration. \method outperforms training-free settings and training-based skill improvement baselines, showing that the framework is effective and not tied to a particular skill source. Ablations show that removing momentum or contrastive diagnosis lowers held-out accuracy, and analysis of batch size, iteration budget, and token cost clarify the behavior and budget of the framework.

In summary, our contributions are three-fold:
\begin{itemize}

    \item We formulate agent skill improvement as optimization over a structured skill artifact, with explicit analogues of parameter, loss evidence, gradient, momentum, and update.
    \vspace{-5pt}
    \item Based on the formulation, we propose \method, a multi-agent framework that diagnoses executions, accumulates recurring patterns, and applies layer-aware skill patches.
    \vspace{-5pt}
    \item Empirical experiments demonstrate that \method improves agents on spreadsheet tasks from both LLM-generated and third-party initial skills, with gains under in-domain and out-of-domain evaluations.

\end{itemize}

\begin{table}[t]
\centering
\small
\setlength{\tabcolsep}{4pt}
\renewcommand{\arraystretch}{1.12}
\begin{tabular}{@{}p{0.42\linewidth}p{0.50\linewidth}@{}}
\toprule
\textbf{Gradient descent} & \textbf{\method optimization} \\
\midrule
Parameter & Structured skill package $S_t$ \\ \addlinespace
Loss evidence & Task outcome and trajectory evidence $\mathcal{E}_t(\mathcal{T}_i)$ \\ \addlinespace
Gradient signal & Task diagnosis $d_{t,i}$ \\ \addlinespace
Momentum & Pattern memory $M_t$ and overlay $O_t$ \\ \addlinespace
Parameter Update & Layer-aware patch $S_{t+1}=\mathrm{Patch}(S_t,\{d_{t,i}\}_{i=1}^{B},M_t,O_t)$ \\
\bottomrule
\end{tabular}
\caption{Conceptual correspondence between gradient descent and \method.}
\label{tab:analogy}
\vspace{-10pt}
\end{table}

\section{Related Work}
\label{sec:related_work}

Recent work has explored agent skills as reusable artifacts that can be generated, updated, and reused by LLM agents. The training-based baselines in our experiments are EvoSkill and Trace2Skill because both consume task executions and produce standalone skill artifacts, which allows all methods to be evaluated under the same initialization, training tasks, backbone model, and held-out split. EvoSkill \citep{alzubi2026evoskill} follows an iterative skill evolution formulation. It analyzes failed executions, turns the resulting diagnoses into new or revised skills, and selects candidates using held-out validation performance. This corresponds to a \textit{failure-driven} update strategy with validation-based selection, while \method optimizes one current skill artifact using both failed executions and contrastive successful executions as loss evidence. Trace2Skill \citep{ni2026trace2skill} follows a trajectory-to-skill distillation formulation. It analyzes a pool of execution trajectories, extracts local lessons, and hierarchically consolidates them into a unified skill directory. This gives an offline trace-distillation strategy, while in comparison, \method repeatedly executes the current skill so that each update changes the evidence observed in later iterations.

Other skill-oriented systems study broader forms of skill acquisition, memory, and reuse. SkillX \citep{wang2026skillx} constructs a plug-and-play skill knowledge base by organizing experience into multi-level skills, refining them with execution feedback, and expanding the library with newly generated skills. SkillClaw \citep{ma2026skillclaw} studies collective skill evolution in multi-user agent ecosystems, where trajectories accumulated across users are aggregated to refine existing skills or extend a shared skill repository. Memento-Skills \citep{zhou2026memento} treats structured markdown skills as persistent memory, enabling agents to retrieve, update, and expand task-specific skills through a read--write learning loop. CoEvoSkills \citep{zhang2026evoskills} constructs complex multi-file skill packages through a skill generator and a separate verifier that critiques executions and provides feedback for later revisions. AutoSkill \citep{yang2026autoskill} focuses on lifelong personalized agents by deriving, maintaining, and reusing skills from user dialogue and interaction traces. Together, these works broaden agent skill learning toward skill libraries, shared repositories, verifier-guided construction, and lifelong personalization.

\begin{figure*}[ht]
\centering
\includegraphics[width=\linewidth]{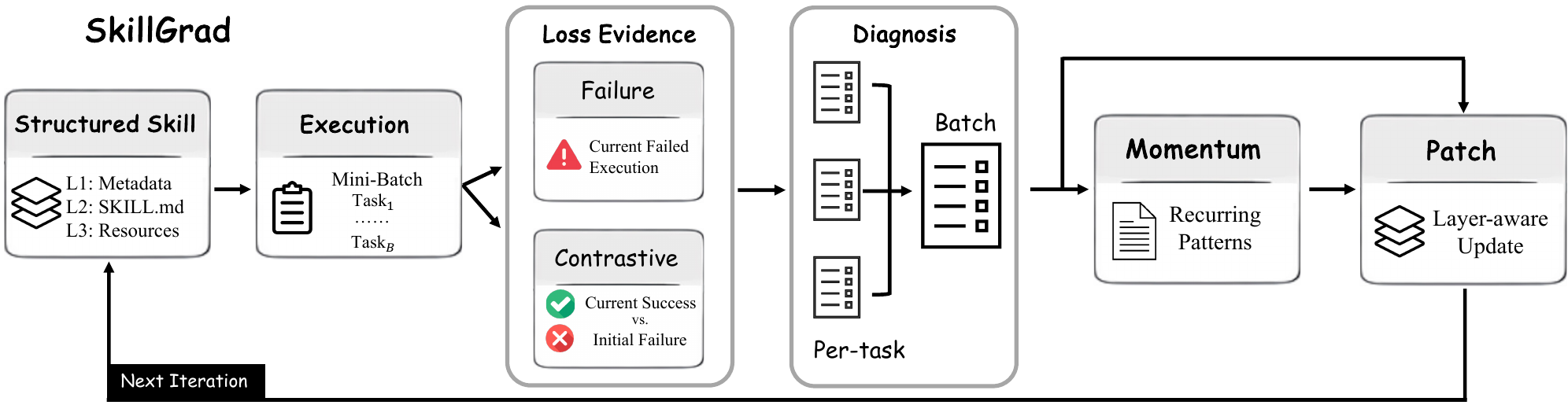}
\caption{Overview of \method. Given a structured skill and a mini-batch of training tasks, the executor produces trajectories and outcomes. \method turns failed executions and contrastive successful executions into loss evidence, diagnoses reusable update signals, accumulates recurring patterns through textual momentum, and applies a layer-aware patch to obtain the next skill.}
\label{fig:skillgrad-overview}
\vspace{-15pt}
\end{figure*}

\section{Methodology}
\label{sec:method}

This section describes how \method instantiates the optimization analogy introduced in Section~\ref{sec:intro}. We first provide the overview of the framework in Section~\ref{subsec:overview}. We then define the skill as the optimizable parameter (Section~\ref{subsec:parameter}), describe execution outcomes and trajectory evidence (Section~\ref{subsec:loss}), construct diagnoses as gradient-like signals (Section~\ref{subsec:grad}), introduce cross-iteration momentum (Section~\ref{subsec:momentum}), and close with the layer-aware skill update (Section~\ref{subsec:parameter_update}).

\subsection{Overview}
\label{subsec:overview}

\method optimizes a structured skill through the iterative loop shown in Figure~\ref{fig:skillgrad-overview}. At each iteration, the executor applies the current skill to a mini-batch of training tasks. The resulting outcomes and trajectories provide loss evidence. Failed executions reveal missing or incorrect guidance, while contrastive successful executions reveal behaviors that the current skill has learned to perform and should preserve. A diagnoser converts these observations into task-level diagnoses and aggregates them into a batch diagnosis. The momentum agent then updates a persistent record of recurring patterns and writes a compact overlay for the current batch. Finally, the patcher edits the structured skill package, producing the skill used in the next iteration. This loop mirrors the operational structure of gradient descent, which evaluates the current parameter, derives updated evidence from observed outcomes, accumulates recurring directions, and applies an update to obtain the next parameter. The correspondence is conceptual, but it gives each module a clear role in the skill optimization process.

\subsection{Parameter}
\label{subsec:parameter}

The trainable parameter in gradient descent is typically the weight vector $\theta$ of a model $f_\theta$. In skill optimization, the parameter is instead the skill package $S$. A skill is not a flat prompt, but a progressively disclosed artifact with three layers:
\begin{itemize}
\vspace{-5pt}
    \item \textbf{L1: Metadata.} YAML skill description.
    \vspace{-5pt}
    \item \textbf{L2: \texttt{SKILL.md} file.} The full body of the \texttt{SKILL.md} file, which contains principles, procedural workflows, operations, code examples, and common pitfalls.
    \vspace{-5pt}
    \item \textbf{L3: Resources.} Additional files form the third layer of the skill, serving as conditional resources for longer procedures, edge cases, and worked examples.
    \vspace{-5pt}
\end{itemize}
We write this as
\begin{equation*}
S = (H, B, \mathcal{Q}),
\end{equation*}
where $H$ denotes the metadata header, $B$ denotes the always-loaded \texttt{SKILL.md} body, and
$\mathcal{Q}$ denotes the set of conditionally loaded resources.

This structure makes skill optimization different from ordinary prompt optimization \citep{agrawal2025gepa, ren2026polca}. Since $B$ is always loaded after the skill activates, it should contain compact and broadly useful guidance. Since $\mathcal{Q}$ is loaded only when referenced, it can contain longer procedures, edge cases, and worked examples without burdening unrelated executions. A useful update must therefore decide not only what knowledge to add, but also where that knowledge should live. Placing narrow task detail in $B$ can distract the executor on future tasks, while placing core workflow guidance only in $\mathcal{Q}$ can prevent
the executor from loading it when needed. \method treats this routing decision as part of the parameter update. Appendix~\ref{appendix:skill_artifact} shows representative initial and optimized skill excerpts, and Appendix~\ref{appendix:l3_usage} summarizes runtime retrieval of learned L3 resources.

\subsection{Loss Evidence}
\label{subsec:loss}

For skill optimization in an agentic setting, the most immediate analogue of a loss is the evaluated
task outcome. Given a task $\mathcal{T}$ and a skill $S$, the executor produces an output that is evaluated against the reference answer. We denote the task success indicator as
\begin{equation*}
\mathcal{R}(S; \mathcal{T}) =
\begin{cases}
1, & \text{if the agent output passes}, \\
0, & \text{otherwise}.
\end{cases}
\end{equation*}
This gives the terminal binary loss
\begin{equation*}
\mathcal{L}_{\mathrm{bin}}(S; \mathcal{T}) =
1 - \mathcal{R}(S; \mathcal{T}).
\end{equation*}

The binary loss is the scalar signal used to evaluate task success, but it is too sparse to be the only signal used to update a structured skill. If updates were based only on $\mathcal{L}_{\mathrm{bin}}$, failed executions would be retained as repair evidence, while every successful execution would collapse to zero and be discarded. This mirrors the limitation of a hard 0-1 loss in supervised learning. A classifier can predict the correct label while still receiving a nonzero cross-entropy loss, because the predictive distribution may not yet be robust. Therefore, gradient descent does not discard a training instance simply because its discrete prediction is correct. Although agent trajectories are not differentiable probability vectors, final correctness likewise does not imply that an execution contains no useful learning signal.

\method therefore uses the binary loss as the terminal evaluation signal, while constructing a richer trajectory-level object as loss evidence. Let
$r_t(\mathcal{T})=\mathcal{R}(S_t;\mathcal{T})$, and let $c_t(\mathcal{T})$ denote the evaluator feedback for the execution at iteration $t$, such as the comparison between the produced output and the reference answer. We intentionally sample the training tasks from failures of the initial skill. Thus, a current success can be paired with the corresponding initial failure. We define the loss evidence as
\begin{equation}
\label{eq:loss-evidence}
\mathcal{E}_t(\mathcal{T}) =
\begin{cases}
\{\tau_t^{-}, c_t\}, & r_t(\mathcal{T}) = 0, \\[4pt]
\{\tau_t^{+}, \tau_0^{-}, c_0\}, & r_t(\mathcal{T}) = 1,
\end{cases}
\end{equation}
where $\tau_t^{-}$ and $\tau_t^{+}$ denote failed and successful trajectories under the current skill $S_t$, and $\tau_0^{-}$ denotes the failed trajectory from the initial skill $S_0$ on the same task. The two branches provide complementary evidence. Failed trajectories identify behaviors associated with high terminal loss and support corrective diagnoses. Contrastive successful trajectories identify what changed between an earlier failure and the current successful execution, such as a more robust coding strategy, a complete inspection step, or a verification step that prevents common errors.

This evidence design distinguishes \method from failure-only skill evolution methods such as \citet{alzubi2026evoskill}. Such methods diagnose failed trajectories, but do not use successful trajectories as diagnostic evidence. In contrast, our framework is motivated by the same intuition as gradient descent that correct terminal outcomes do not imply zero learning signal. Even when the current skill succeeds on a task, the successful trajectory can still provide useful information when contrasted with nearby failures. Therefore, the loss design preserves both negative and positive evidence, enabling more informative and principled skill optimization.

\subsection{Gradient Signals}
\label{subsec:grad}

In gradient-based optimization, the loss becomes actionable through the gradient, which indicates a local direction that relates the observed error to a change in the parameters. Given parameters $\theta_t$ and a training example $x_i$, the gradient $\nabla_\theta \mathcal{L}(\theta_t; x_i)$ provides a per-example update signal. For a mini-batch $\mathcal{B}$, the optimization signal is aggregated across samples:
\begin{equation*}
g_t =
\frac{1}{|\mathcal{B}|}
\sum_{x_i \in \mathcal{B}}
\nabla_{\theta}\mathcal{L}(\theta_t; x_i).
\end{equation*}

For a structured skill, no numeric derivative is available. The parameter is a natural-language file package, and the executor's behavior depends on tool use, intermediate reasoning, and external files. \method therefore constructs a textual counterpart of gradients through diagnosis. For each task $\mathcal{T}_{t,i}$ in the mini-batch, let
$e_{t,i}=\mathcal{E}_t(\mathcal{T}_{t,i})$ be the loss evidence defined in Eq.~\ref{eq:loss-evidence}. The diagnoser has access to the current skill, the task, and this evidence, and produces
\begin{equation*}
d_{t,i} =
\mathrm{Diag}(S_t, \mathcal{T}_{t,i}, e_{t,i}).
\end{equation*}

A diagnosis is not a score or a summary of the final answer, but an evidence-grounded update signal. It identifies which execution behavior the evidence points to as responsible for the outcome and describes what reusable behavior should be repaired or preserved. For failed trajectories, it explains why the produced output differs from the ground truth and what general behavior would have avoided the error. For contrastive successful trajectories, it explains what changed relative to the earlier failure and whether the successful behavior is reusable. Conditioning the diagnosis on
$S_t$ is important because the same execution evidence can imply different updates depending on whether the relevant guidance is absent from the skill, present but too weak, or already present but ignored by the executor.

Following the mini-batch structure of gradient descent, \method obtains one diagnosis for each task and collects them into a batch-level diagnosis set:
\begin{equation*}
D_t = \{d_{t,1}, d_{t,2}, \ldots, d_{t,B}\}.
\end{equation*}
The textual diagnoses cannot be averaged as vectors in mini-batch gradient descent. Thus, we follow \citet{yuksekgonul2024textgrad} to preserve the per-task signals. The next stage, the momentum mechanism, performs the semantic aggregation. It identifies which diagnosed mechanisms are new, recurring, already covered, or still unresolved, and then passes that state to the patcher.

\subsection{Momentum}
\label{subsec:momentum}

In gradient descent with momentum, the optimizer maintains a persistent state that accumulates past update directions:
\begin{equation*}
v_t = \gamma v_{t-1} + g_t,
\end{equation*}
where $g_t$ is the current batch gradient and $v_t$ is the momentum vector. The purpose of this state is not only to remember the latest gradient, but to stabilize updates by reinforcing directions that recur across iterations.

\method introduces an analogous textual momentum mechanism. Unlike numeric momentum, textual momentum does not perform arithmetic accumulation or decay. It implements the optimizer-state role by tracking recurring semantic directions and their absorption status. Specifically, the momentum agent maintains a persistent pattern memory:
\begin{equation*}
M_t, O_t = \mathrm{Momentum}(M_{t-1}, D_t, S_t),
\end{equation*}
where $M_t$ stores cross-iteration patterns and $O_t$ is a compact overlay for the current patch. The memory records reusable mechanisms that have appeared in past diagnoses, such as a missing workbook-inspection step, a wrong lookup direction, a fragile formula choice, or a verification behavior that repeatedly enables success. Each pattern is associated with the evidence that supports it and with the part of the skill that currently covers or fails to cover it.

The momentum stage serves three roles. First, it performs semantic accumulation. Multiple task diagnoses that express the same underlying mechanism are treated as one recurring update direction rather than independent one-off patches. Second, it conditions the update on the current skill state. A recurring pattern should lead to different patches depending on whether the skill lacks it, states it too vaguely, or already contains adequate guidance that should be preserved. This reduces update churn and helps stabilize the optimized artifact. Third, it carries successful contrastive behaviors forward. This prevents the patcher from only chasing failures and helps preserve behaviors that newly solved tasks reveal as useful.

Thus, textual momentum plays an analogous optimizer-state role as optimizer momentum, converting noisy, local per-example signals into a more stable update context. The conceptual correspondence is operationally important. With momentum, the patcher sees whether a pattern is new, recurring, unresolved, or already absorbed into the skill, making each update less dependent on the current batch alone.

\subsection{Parameter Updates}
\label{subsec:parameter_update}

The final step of each iteration is the parameter update. In gradient descent, the update applies the optimizer state to the parameter vector:
\begin{equation*}
\theta_{t+1} = \theta_t - \eta v_t.
\end{equation*}
In \method, the patcher agent applies the textual optimizer state to the structured skill:
\begin{equation*}
S_{t+1} = \mathrm{Patch}(S_t, D_t, M_t, O_t).
\end{equation*}
The patcher reads the current skill $S_t$, the task-level diagnoses $D_t$, the persistent memory $M_t$, and the current overlay $O_t$. These inputs have complementary roles. $D_t$ preserves the raw per-example update signals, $M_t$ records whether a mechanism is recurring or already handled, and $O_t$ focuses the current edit on the patterns that should be considered in this iteration.

The key design choice is that the patcher updates patterns, not tasks. If several diagnoses point to the same mechanism, the patcher produces one generalized edit rather than a list of task-specific fixes. This mirrors the role of a mini-batch update, where multiple examples jointly determine one parameter change, and prevents the skill from becoming an append-only record of the training set.

The update is also layer-aware. Since $S_t=(H_t,B_t,\mathcal{Q}_t)$ is a structured parameter, the patcher must decide both what behavior should change and where the change belongs in the skill hierarchy. This is the key difference from optimizing a flat prompt: the learned content must remain organized so that future executions can retrieve and apply it under the appropriate conditions.

After the patch, the edited skill becomes the parameter for the next execution batch. This closes the optimization loop where each update changes the executor's future behavior distribution, in turn changing the loss evidence and diagnoses observed in later iterations.

\section{Experiments}
\label{sec:exp}

\subsection{Experimental Settings}
\label{subsec:exp_setting}

\begin{table*}[ht]
\vspace{-10pt}
\centering
\small
\setlength{\tabcolsep}{5pt}
\renewcommand{\arraystretch}{1.15}
\begin{tabular}{@{}llcccc@{}}
\toprule
\multirow{2}{*}{\textbf{Init.}} 
& \multirow{2}{*}{\textbf{Method}} 
& \multicolumn{2}{c}{\textbf{GPT-5.4}} 
& \multicolumn{2}{c}{\textbf{GPT-4.1}} \\
\cmidrule(lr){3-4} \cmidrule(lr){5-6}
& & \textbf{SpreadsheetBench} 
& \textbf{WikiTQ (OOD)} 
& \textbf{SpreadsheetBench} 
& \textbf{WikiTQ (OOD)} \\
\midrule
\multicolumn{6}{@{}l}{\textit{Training-free}} \\
\hdashline
-- & No Skill 
& 62.50 
& 78.57
& 44.17 
& 52.86 \\
LLM-gen. & Base \texttt{xlsx} Skill 
& 55.83
& 77.14 
& 36.67 
& 48.58 \\
Third-party & Base \texttt{xlsx} Skill
& 60.00
& 78.57
& 33.33
& 42.86 \\
\midrule
\multicolumn{6}{@{}l}{\textit{Training-based skill improvement}} \\
\hdashline
\multirow{3}{*}{LLM-gen.} & Trace2Skill 
& 65.28 \scorestd{2.93}
& 79.05 \scorestd{1.35} 
& 37.22 \scorestd{3.54} 
& 60.00 \scorestd{5.83} \\
 & EvoSkill 
& 68.06 \scorestd{0.48}
& 78.09 \scorestd{0.67} 
& 37.22 \scorestd{2.58} 
& 53.33 \scorestd{7.50} \\
& \textbf{\method (Ours)} 
& \textbf{71.11} \scorestd{1.73}
& \textbf{82.38} \scorestd{1.78} 
& \textbf{54.17} \scorestd{3.54}
& \textbf{73.65} \scorestd{2.50} \\
\addlinespace[2pt]
\multirow{3}{*}{Third-party} & Trace2Skill
& 63.89 \scorestd{0.48}
& 81.91 \scorestd{1.78}
& 38.89 \scorestd{3.85}
& 51.43 \scorestd{1.17} \\
& EvoSkill
& 63.61 \scorestd{1.92}
& 80.95 \scorestd{3.56}
& 36.94 \scorestd{1.92}
& 43.33 \scorestd{4.86} \\
& \textbf{\method (Ours)}
& \textbf{69.44} \scorestd{0.48}
& \textbf{83.34} \scorestd{0.67}
& \textbf{45.83} \scorestd{3.66}
& \textbf{53.81} \scorestd{2.43} \\
\bottomrule
\end{tabular}
\caption{
Main experimental results on SpreadsheetBench Verified and WikiTableQuestions. For training-based methods, we report mean with standard deviation over three random seeds used for training-set selection. Results on both benchmarks are reported as accuracy (\%). ``LLM-gen.'' denotes the default LLM-generated \texttt{xlsx} initialization, and ``Third-party'' denotes the third-party downloaded \texttt{xlsx} initialization. The training-based skill improvement block includes EvoSkill and Trace2Skill adapted to the same fixed-training setting, together with \method. Best means within each initialization block are shown in \textbf{bold}.
}
\label{tab:main-results}
\end{table*}

\noindent\textbf{Datasets and Evaluation.} We evaluate domain-specific agent skills primarily on \textsc{SpreadsheetBench Verified}, a human-validated subset of SpreadsheetBench designed for reliable automatic evaluation \citep{ma2024spreadsheetbench}. The benchmark is derived from real-world Excel forum questions and covers both cell-level and sheet-level spreadsheet manipulation. Following the official evaluation protocol, the agent-generated executable solution is applied to the input workbook, formulas are recalculated, and the resulting workbook is compared with the golden workbook within annotated answer ranges. A task is considered correct only if all required output cells match the ground truth; we report the percentage of fully solved tasks. For out-of-domain transfer, we additionally evaluate on \textsc{WikiTableQuestions} (WikiTQ), a semi-structured table question answering benchmark over Wikipedia tables \citep{pasupat2015compositional}. Following the official WikiTQ evaluation protocol, we compare the predicted answer denotation against the gold denotation and report accuracy. Due to the high cost of agent-based execution, we use fixed sampled subsets for evaluation and provide more details in Appendix~\ref{appendix:config}.

\noindent\textbf{Models.} We evaluate two backbone LLMs, gpt-5.4 and gpt-4.1, as the reasoning engine for all agent variants. Each reported run uses a single backbone throughout the full framework, meaning that the gpt-5.4 run uses gpt-5.4 for all agents, and the gpt-4.1 run uses gpt-4.1 for all agents, with no cross-model mixing.

\noindent\textbf{Baselines.} We include training-free settings to measure the effect of using no skill or directly using the initialization skill. For methods involving training, EvoSkill \citep{alzubi2026evoskill} is formulated as skill evolution through iterative failure analysis and validation selection, while Trace2Skill \citep{ni2026trace2skill} is formulated as trajectory-to-skill distillation over frozen initial skill executions. Both methods can produce an updated skill artifact from task executions, so we adapt them to our fixed-training setting and compare all training-based methods under the same initialization skill, selected training tasks, backbone model, and held-out evaluation split.

\noindent\textbf{Training Configurations.} Since we focus on optimizing a given agent skill, training-based methods start from the same skill initialization within each block of Table~\ref{tab:main-results}. The default setting uses an LLM-generated base \texttt{xlsx} skill. We additionally test a third-party \texttt{xlsx} skill to evaluate whether \method can improve an externally sourced skill package. For each run, we randomly select 40 SpreadsheetBench Verified tasks as the training set and evaluate the optimized skill on a fixed held-out test split. We set the batch size to 4 and train for 10 iterations. This configuration is chosen due to the heavy computational overhead of an LLM-agent forward pass, where each example requires a full execution trajectory, and it lets one optimization run cover the 40-task training set once. Additional configuration details are provided in Appendix~\ref{appendix:config}.

\subsection{Results}
\label{subsec:res}

Table~\ref{tab:main-results} shows that \method consistently outperforms both training-free and training-based baselines. With the LLM-generated initial skill on SpreadsheetBench, \method reaches $71.11\%$ using gpt-5.4, improving over the average of the two training-based baselines by $4.44$ percentage points (pp). The pattern is even stronger with gpt-4.1. \method reaches $54.17\%$, while Trace2Skill and EvoSkill tie at $37.22\%$. This suggests that the benefit is not tied to a single high-performing backbone. The optimization framework can also help a weaker executor recover useful domain behavior.

The training-free rows further motivate optimizing skills rather than merely generating them once. The LLM-generated base \texttt{xlsx} skill hurts performance compared with the no-skill setting on both backbones, dropping from $62.50\%$ to $55.83\%$ with gpt-5.4 and from $44.17\%$ to $36.67\%$ with gpt-4.1. This agrees with prior observations that automatically generated skills can be noisy or misleading \citep{li2026skillsbench}. \method reverses this degradation by treating the skill as an optimizable artifact: starting from the same base skill, it improves over the base skill by $15.28$ pp with gpt-5.4 and $17.50$ pp with gpt-4.1.

The third-party \texttt{xlsx} skill shows the same trend. On SpreadsheetBench, \method improves the base skill from $60.00\%$ to $69.44\%$ with gpt-5.4 and from $33.33\%$ to $45.83\%$ with gpt-4.1. On WikiTQ, it improves the same initialization from $78.57\%$ to $83.34\%$ with gpt-5.4 and from $42.86\%$ to $53.81\%$ with gpt-4.1. These results indicate that \method is not specific to the particular LLM-generated skill used in the default setting.

The WikiTQ results test whether the optimized spreadsheet skill overfits to SpreadsheetBench-style tasks. Although WikiTQ differs in task format and output space, \method still obtains the best OOD accuracy in every initialization and backbone block. With the LLM-generated initialization, \method improves over the average of Trace2Skill and EvoSkill by $3.81$ pp using gpt-5.4. With gpt-4.1, the gain is larger, improving over Trace2Skill by $13.65$ pp and over EvoSkill by $20.32$ pp. These results provide evidence that the optimized skills contain reusable procedural guidance beyond the training tasks.

\subsection{Ablations}
\label{subsec:ablations}

We ablate two components of \method, the cross-iteration momentum and the contrastive diagnoses from tasks that previously failed but now succeed. Unless otherwise stated, component ablations use gpt-5.4 with the same training configuration and a fixed training-set seed. We use the same evaluation protocol and metric, reporting results on the same SpreadsheetBench Verified test split as the main experiments.

\begin{table}[t]
\centering
\small
\setlength{\tabcolsep}{4pt}
\renewcommand{\arraystretch}{1.12}
\begin{tabular}{@{}lcc@{}}
\toprule
\textbf{Variant} & \textbf{Acc.} & \textbf{$\Delta$ Acc.} \\
\midrule
Full \method & \textbf{72.50} & -- \\
No momentum & 65.83 & $-6.67$ \\
Failure-only diagnosis & 68.33 & $-4.17$ \\
\bottomrule
\end{tabular}
\caption{Ablation results on the SpreadsheetBench Verified test split. ``$\Delta$ Acc.'' reports the absolute change in accuracy, in percentage points, relative to the full \method run under the same training seed and test split.}
\label{tab:component-ablation}
\end{table}

Table~\ref{tab:component-ablation} shows that removing momentum reduces held-out accuracy from $72.50\%$ to $65.83\%$, while removing contrastive diagnosis decreases accuracy to $68.33\%$. Both ablated variants still receive training trajectories and edit a skill, and their scores remain in the range of the adapted training-based baselines in Table~\ref{tab:main-results}. The remaining gap to the full method indicates that the complete update loop benefits from carrying recurring evidence across iterations and retaining successful recoveries as update evidence. Appendix~\ref{appendix:ablation_artifacts} provides artifact-level analysis of how these ablations differ from the full method.

\section{Analysis}
\label{sec:analy}

In this section, we analyze hyperparameters in \method and trace the training cost. We focus on SpreadsheetBench Verified with gpt-5.4, where the main component ablations and training logs were collected. More qualitative analyses are provided in Appendix~\ref{appendix:dynamics}.

\subsection{Optimization Behavior}
\label{subsec:analy_opt_ablation}

\begin{figure}[t]
\centering
\begin{subfigure}[t]{0.48\linewidth}
    \centering
    \includegraphics[width=\linewidth]{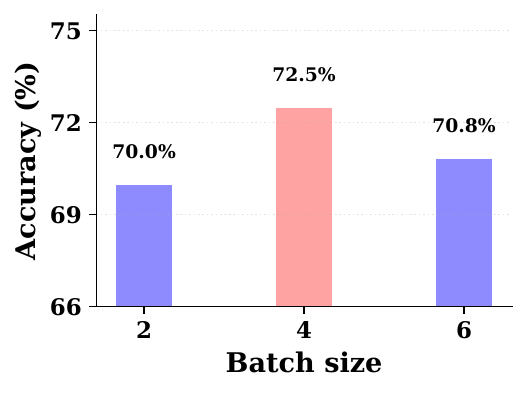}
    \caption{Batch size}
    \label{fig:batch_size_ablation}
\end{subfigure}
\hfill
\begin{subfigure}[t]{0.48\linewidth}
    \centering
    \includegraphics[width=\linewidth]{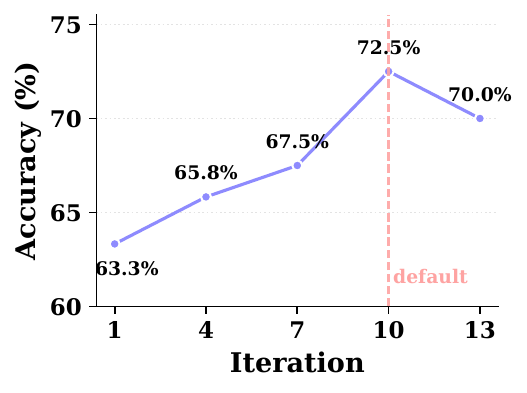}
    \caption{Training iteration}
    \label{fig:iteration_ablation}
\end{subfigure}
\caption{
Hyperparameter analysis on the held-out SpreadsheetBench Verified test set with gpt-5.4.}
\label{fig:optimization_ablations}
\end{figure}

\noindent\textbf{Batch size.} Figure~\ref{fig:optimization_ablations}(a) varies the mini-batch size while holding the number of textual updates fixed at $10$. With batch size $2$,
\method sees $20$ training trajectories and reaches $70.0\%$ accuracy. The default batch size $4$ corresponds to one pass over the $40$-task training pool and reaches $72.5\%$. Increasing the batch size to $6$ gives the optimizer $60$ trajectory slots, including repeated tasks after the first pass, and reaches $70.8\%$.

This pattern suggests that the method is reasonably robust to the batch sizes tested, but that larger batches do not automatically improve the optimized skill. Small batches under-sample the training pool under the fixed update budget. Larger batches provide more evidence per update, but they also require one textual patch to compress a wider set of diagnoses into a single skill edit. In this setting, batch size $4$ gives the best observed balance between evidence per update and update frequency.

\noindent\textbf{Iteration budget.}
Figure~\ref{fig:optimization_ablations}(b) evaluates checkpoints from the same training trajectory. Accuracy rises from $63.3\%$ at iteration~$1$ to $65.8\%$ at iteration~$4$, $67.5\%$ at iteration~$7$, and $72.5\%$ at the default iteration~$10$. Continuing training for three additional iterations gives $70.0\%$ at iteration~$13$.

The checkpoint curve supports the fixed-budget choice used in the main experiments. Accuracy improves as more mini-batches are processed and reaches the best observed checkpoint at the one-pass setting. The decline after iteration~$10$ also shows that textual updates are not monotonic. Additional edits can begin to trade off against earlier general rules, so we report the fixed iteration budget rather than selecting the best checkpoint after evaluation.

\subsection{Training Cost}
\label{subsec:analy_cost}

\begin{figure}[t]
\centering
\begin{subfigure}[t]{0.48\linewidth}
    \centering
    \includegraphics[width=\linewidth]{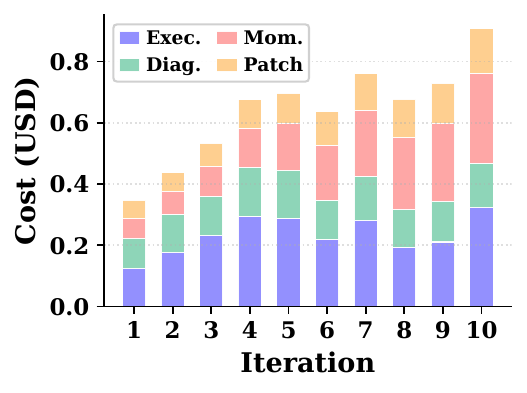}
    \caption{Per-iteration cost}
    \label{fig:cost_per_iter}
\end{subfigure}
\hfill
\begin{subfigure}[t]{0.48\linewidth}
    \centering
    \includegraphics[width=\linewidth]{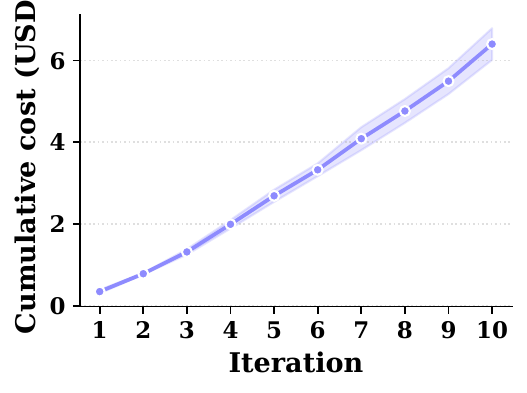}
    \caption{Cumulative cost}
    \label{fig:cost_cumulative}
\end{subfigure}
\caption{
Training cost analysis for \method with gpt-5.4, averaged over three seeds.}
\label{fig:cost}
\end{figure}

We record API cost at each stage of every training iteration. Figure~\ref{fig:cost} shows that a full ten-iteration \method run costs
USD~$6.40 \pm 0.38$ on average across three gpt-5.4 training seeds. The per-iteration cost increases from about USD~$0.35$ at iteration~$1$ to about USD~$0.85$ at iteration~$10$.

The cost grows gradually over training. Execution and diagnosis stay roughly flat because the mini-batch size is fixed. The increasing part comes mainly from the momentum and patch stages, whose prompts include improvement of skill and the accumulated pattern state. We report this accounting to make the optimization budget explicit. \method produces a reusable skill artifact without fine-tuning model weights or relying on a large validation sweep over many candidate skills.

\section{Conclusion}
\label{sec:conclusion}

We present \method, an optimization-inspired framework for agent skill improvement. \method casts a structured skill package as the optimizable artifact and maps execution evidence, diagnosis, momentum, and layer-aware patching to the main stages of an optimization loop. Across SpreadsheetBench Verified and WikiTableQuestions, \method improves skills initialized from both LLM-generated and third-party sources. Empirical results suggest a practical path for improving agent skills through structured optimization.

\newpage
\section*{Limitations}

\method is evaluated primarily on spreadsheet-centered tasks, with WikiTableQuestions used as an out-of-domain transfer setting. Future work can test whether the same optimization-inspired framework transfers to other skill domains such as web automation, document editing, and codebase maintenance. In addition, our current analysis is empirical and qualitative. A more formal account of when textual diagnoses and momentum states yield stable skill updates would further strengthen the connection between agent skills optimization and classical optimization.

\bibliography{custom}

\appendix

\section{Experimental Details}
\label{appendix:exp_detail}

In this appendix, we summarize the experimental protocol used in Section~\ref{sec:exp}. Unless otherwise specified, all SpreadsheetBench experiments use the same fixed split and evaluation procedure.

\subsection{Configurations}
\label{appendix:config}

\noindent\textbf{SpreadsheetBench split.}
We construct one canonical split from SpreadsheetBench Verified and use it throughout the paper. We first sample 200 tasks as the training candidate pool and reserve another 200 tasks as held-out tasks. The held-out tasks are shuffled once with a fixed seed. The first 20 tasks form the validation set used by EvoSkill, and the next 120 tasks form the fixed test set used for all reported SpreadsheetBench results. The remaining held-out tasks are kept unused. This protocol keeps training, validation, and test tasks disjoint while avoiding repeated resampling of the test set across methods.

\begin{table}[ht]
\centering
\small
\setlength{\tabcolsep}{4pt}
\begin{tabularx}{\linewidth}{@{}l c X@{}}
\toprule
\textbf{Split} & \textbf{Size} & \textbf{Usage} \\
\midrule
Training pool & 200 & Pool for training task selection \\
Validation & 20 & Skill selection for EvoSkill \\
Test & 120 & SpreadsheetBench evaluation \\
Unused held-out & 60 & Reserved and not used in this paper \\
\bottomrule
\end{tabularx}
\caption{SpreadsheetBench split used in the experiments.}
\label{tab:appendix-split}
\end{table}

\noindent\textbf{WikiTableQuestions subset.}
For out-of-domain evaluation, we randomly sample 70 examples from the original WikiTableQuestions dataset and keep this subset fixed for all methods and backbone models. We use this subset only for evaluation.

\noindent\textbf{Training task selection.}
For each backbone model and skill initialization, we first execute the initialization skill on the 200-task candidate pool. We then collect tasks that the initialization fails to solve and sample 40 of them as the training set for each training seed. The main experiments use training seeds 0, 1, and 2. This makes the optimization problem nontrivial while keeping the training budget small enough for agent-based execution. Table~\ref{tab:appendix-failure-pools} reports the failure pools used to sample training tasks.

\begin{table}[t]
\centering
\small
\setlength{\tabcolsep}{4pt}
\begin{tabular}{@{}l l r r@{}}
\toprule
\textbf{Initialization} & \textbf{Model} & \textbf{Fail} & \textbf{Solve} \\
\midrule
LLM-generated & GPT-5.4 & 76 & 124 \\
LLM-generated & GPT-4.1 & 117 & 83 \\
Third-party & GPT-5.4 & 72 & 128 \\
Third-party & GPT-4.1 & 124 & 76 \\
\bottomrule
\end{tabular}
\caption{Failure pools on the 200-task SpreadsheetBench training candidates.}
\label{tab:appendix-failure-pools}
\end{table}

\noindent\textbf{Optimization settings.}
For SkillGrad, we use a batch size of 4 and train for 10 iterations unless an ablation explicitly changes this setting. With 40 training tasks, the default setting corresponds to one pass over the entire sampled training set. Each training example is executed with a maximum of 30 agent turns. We do not use a validation set for SkillGrad to select the best-performed skill or to determine whether to discard the optimization after each iteration, out of consideration for cost and latency. The final skill after 10 iterations is evaluated on the fixed 120-task test set.

\noindent\textbf{Evaluation.}
For SpreadsheetBench, the agent produces an output workbook. The evaluator recalculates spreadsheet formulas and compares the annotated answer ranges against the golden workbook. We follow the standard evaluation to report hard accuracy, where a task is counted as correct only when all required cells match. For WikiTableQuestions, all methods are evaluated on the same fixed sampled subset, and we report denotation accuracy following the standard WikiTQ protocol.

\noindent\textbf{Analysis experiments.}
The hyperparameter analysis in Section~\ref{sec:analy} uses the LLM-generated \texttt{xlsx} initialization with GPT-5.4 and training seed 0. The batch-size ablation keeps the number of training tasks fixed at 40 and the number of iterations fixed at 10, and only changes the batch size. The iteration analysis keeps the number of training tasks fixed at 40 and the batch size fixed at 4, and evaluates intermediate or continued checkpoints. The component ablations also use the LLM-generated initialization. 

\subsection{Skill Instantiation}
\label{appendix:skill_inst}

In this work, we focus on optimizing a given agent skill while maintaining the three-layer structure. Our default initialization is an LLM-generated \texttt{xlsx} skill. This initialization contains a concise \texttt{SKILL.md} file and serves as the starting point for the main analysis, hyperparameter studies, and component ablations.

We additionally consider a third-party \texttt{xlsx} skill downloaded from a public repository \footnote{\url{https://github.com/lawve-ai/awesome-legal-skills/tree/main/skills/xlsx-processing-openai}.}. This skill is used only as an alternative initialization in the main results table. It lets us test whether SkillGrad can improve a more comprehensive starting skill, while keeping the rest of the analysis tied to the default LLM-generated initialization. When an optimized skill needs additional resources, all files other than \texttt{SKILL.md} are placed under \texttt{references/*.md}. This simplifies the structure, while keeping the skill format consistent across methods and initializations.

\section{L3 Resource Usage}
\label{appendix:l3_usage}

We count a task as L3-activated when its trajectory contains at least one \texttt{read\_reference} tool call. Table~\ref{tab:l3-usage} reports L3 activation on the held-out SpreadsheetBench Verified test split with gpt-5.4. For \method, we use the three LLM-generated-initialization evaluations that produce the in-domain result in Table~\ref{tab:main-results}. These runs solve 87, 83, and 86 tasks, respectively.

\begin{table}[t]
\centering
\small
\setlength{\tabcolsep}{4pt}
\renewcommand{\arraystretch}{1.12}
\begin{tabular}{@{}lccc@{}}
\toprule
\textbf{Setting} & \textbf{L3 files} & \textbf{L3 reads} & \textbf{Activation} \\
\midrule
No skill & 0 & 0 & 0/120 (0.0) \\
LLM-gen base & 0 & 0 & 0/120 (0.0) \\
Third-party base & 4 & 127 & 113/120 (94.2) \\
\method & 1 $\sim$ 3 & 267 & 259/360 (71.9) \\
\bottomrule
\end{tabular}
\caption{L3 activation on SpreadsheetBench Verified with gpt-5.4.}
\label{tab:l3-usage}
\end{table}

The LLM-generated base skill contains only \texttt{SKILL.md}, so no L3 resources can be retrieved. After optimization, \method creates conditional reference files and the executor reads them on $259$ out of $360$ held-out task-runs. The third-party base skill also contains L3 files, but they are general \texttt{openpyxl} reference files and are read on nearly every task. We therefore interpret L3 activation together with the content of the retrieved resources.

\begin{table*}[t]
\centering
\small
\setlength{\tabcolsep}{4pt}
\renewcommand{\arraystretch}{1.12}
\begin{tabularx}{0.98\textwidth}{@{}llrX@{}}
\toprule
\textbf{Run} & \textbf{Learned L3 resource} & \textbf{Reads} & \textbf{Behavior stored in L3} \\
\midrule
seed 0 & \texttt{mapping\_shapes} & 40 & Structural remaps between source records and destination layouts, including sparse headers, grouped outputs, two-axis lookups, and returned fields from selected records. \\
seed 0 & \texttt{formula\_vs\_python} & 38 & Decisions between live formulas and Python-written final values, with checks for output domains, formula-produced drivers, repeated paired columns, and post-write verification. \\
seed 1 & \texttt{non\_target\_change\_check} & 119 & Verification that targeted edits do not change unrelated workbook cells. \\
seed 2 & \texttt{formula\_compatibility} & 63 & Fallback from fragile formulas to literal values when formulas depend on unsupported functions, dynamic arrays, uncertain recalculation, or mismatched range shapes. \\
seed 2 & \texttt{repeated\_key\_transfers} & 6 & Duplicate-key transfers, repeated source blocks, and contiguous-run logic without accidental aggregation. \\
seed 2 & \texttt{formula\_backed\_cells} & 1 & Safe reading of formula-backed helper cells and spilled ranges. \\
\bottomrule
\end{tabularx}
\caption{Learned L3 resources used by the \method evaluations in Table~\ref{tab:main-results}.}
\label{tab:l3-topics}
\end{table*}

Table~\ref{tab:l3-topics} shows that the learned L3 resources are procedural and task-conditional. Seed~0 and seed~2 place narrower mapping and formula procedures in L3 and retrieve them for about half of the held-out tasks. Seed~1 places a broader non-target-change check in L3 and retrieves it on almost every task. This variation reflects the fact that L3 files are produced independently in each optimization run. Across these runs, the executor retrieves learned L3 resources during evaluation, showing that the extra hierarchy is exercised on held-out tasks.

\section{Qualitative Training Dynamics}
\label{appendix:dynamics}

This section presents qualitative training dynamics of the skill optimization process. \method optimizes discrete text files rather than differentiable parameters, so these diagnostics are descriptive rather than a formal convergence analysis. We use them to examine whether the optimization-inspired loop exhibits interpretable training behavior, as reflected in the changes of the skill structure, the momentum state, and the patch magnitude.

\subsection{Skill Structure}

\begin{figure}[t]
\centering
\begin{subfigure}[t]{0.48\linewidth}
    \centering
    \includegraphics[width=\linewidth]{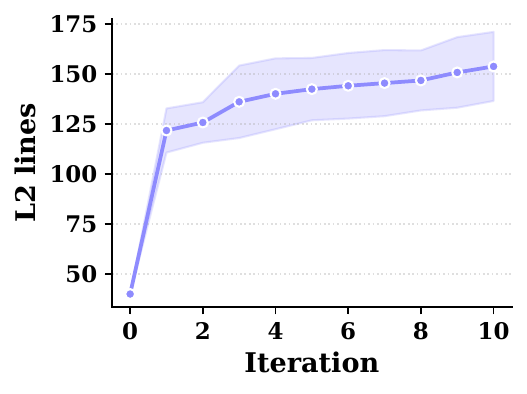}
    \caption{L2 size}
\end{subfigure}
\hfill
\begin{subfigure}[t]{0.48\linewidth}
    \centering
    \includegraphics[width=\linewidth]{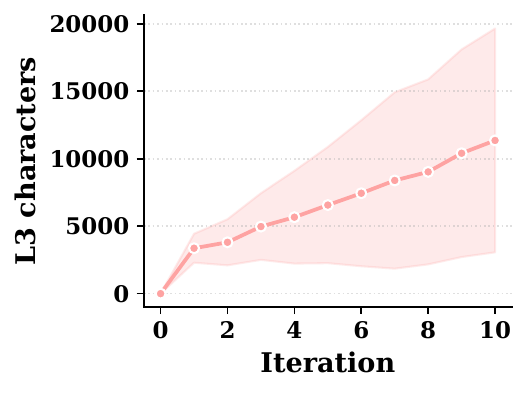}
    \caption{L3 size}
\end{subfigure}
\caption{Skill artifact size across training iterations under \method, averaged over three training seeds with gpt-5.4. Shaded bands denote the per-iteration standard deviation across seeds.}
\label{fig:skill_structure}
\end{figure}

A useful skill should keep broadly reusable guidance in L2 and longer triggerable procedures in L3. Figure~\ref{fig:skill_structure} reports the size of each layer across iterations.

The L2 layer in panel a bootstraps quickly. Iteration~1 expands \texttt{SKILL.md} from the base $40$ lines to roughly $120$ lines, mostly by
introducing workbook classification and mapping shape rules. From iteration~2 onward, the L2 line count stays in a narrow band between roughly $130$ and $155$ lines, with a final mean of $154$ lines. This suggests that later updates mostly refine the always-loaded guidance instead of appending every new lesson to L2.

The L3 layer in panel b behaves differently. It grows almost monotonically from zero to roughly $13$k characters by iteration~10. This is consistent with the role of L3 as conditional storage for algorithmic and edge case material, since each L3 file is loaded only when its L2 pointer is relevant. The contrast between the L2 plateau and the L3 growth illustrates the layer-aware update design in Section~\ref{subsec:parameter}. The added material is routed into different parts of the skill rather than appended uniformly to the always-loaded file.

\subsection{Momentum Dynamics}

\begin{figure}[t]
\centering
\includegraphics[width=0.95\linewidth]{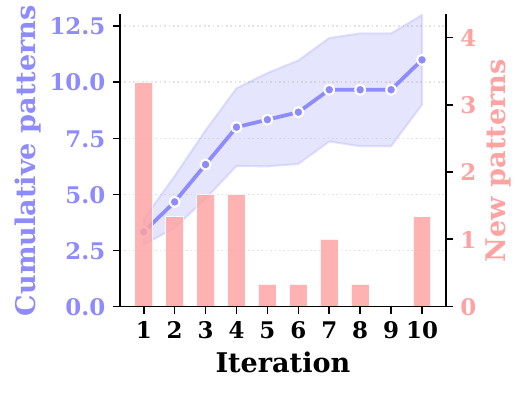}
\caption{Momentum state dynamics across training iterations, averaged over three seeds (gpt-5.4). The purple line and shaded band are the cumulative number of distinct patterns tracked in $M_t$. The red bars are the number of patterns whose first appearance falls in that iteration. The cumulative count saturates near iteration~7, and the new-pattern rate decays from roughly $3$ to under $1$ per iteration.}
\label{fig:momentum_dynamics}
\end{figure}

The momentum agent maintains a persistent record $M_t$ of recurring failure and success patterns, with each pattern listing the iterations where it appeared. We use two derived quantities to characterize this record across iterations. The cumulative pattern count is the size of $M_t$, which measures how many distinct patterns have been tracked by iteration~$t$. The new pattern count is how many entries in $M_t$ list iteration~$t$ as their first appearance.

Figure~\ref{fig:momentum_dynamics} reports both quantities. The cumulative pattern count reaches roughly $10$ by iteration~7 and changes little afterward, even though training continues through iteration~10. As a qualitative signal, this is consistent with a finite set of recurring mechanisms being gradually absorbed into the momentum record. The new pattern rate also decays. Iteration~1 introduces roughly three new patterns on average, while iterations~5~9 introduce zero or one. Later iterations, therefore, mostly reinforce existing update directions instead of opening up new ones. In addition, the active pattern count, measured by the number of patterns whose \texttt{appeared\_in} list contains iteration~$t$, stays in a stable $4 \sim 5$ band from iteration~$3$ onward. Together, these signals align with the intended stabilizing role of momentum. Later updates continue to receive recurring evidence, while the patcher sees each batch in the context of earlier patterns.

\subsection{Patcher Magnitude Behavior}

\begin{figure}[t]
\centering
\includegraphics[width=0.95\linewidth]{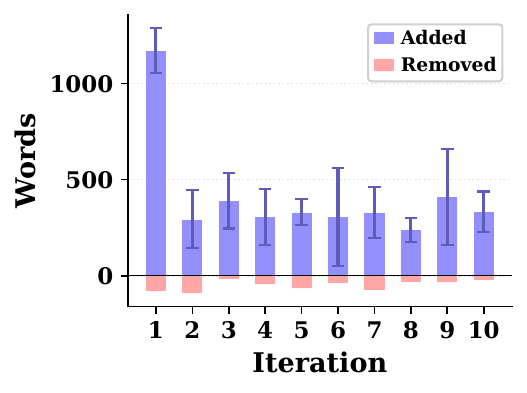}
\caption{Patch magnitude per iteration, measured at the content level by diffing consecutive skill snapshots over the full \texttt{xlsx/} tree. Bars show the mean number of words added (above zero) and removed (below zero) across three seeds, with error bars for added words. Iteration~1 is dominated by the bootstrap patch and adds roughly $1170$ words; iterations~2$\sim$10 average about $330$ added words.}
\label{fig:patch_magnitude}
\end{figure}

Because \method updates text rather than numeric parameters, there is no scalar update magnitude to report. As a descriptive proxy, Figure~\ref{fig:patch_magnitude} reports per-iteration word-level diff totals across the whole \texttt{xlsx/} tree, including \texttt{SKILL.md} and every \texttt{references/*.md}.

Two regimes emerge. Iteration~1 is qualitatively different. It rewrites the base skill substantially, adding roughly $1170$ words and removing about $80$. From iteration~2 onward, the mean added word count drops to about $330$ per iteration and stays in a $240$--$410$ band through iteration~10, while removed words remain consistently low. This pattern is a useful sanity check for the training loop. After the initial bootstrap edit, later patches are smaller and more incremental, rather than repeatedly rewriting the whole skill. This behavior is consistent with evidence-grounded editing under the L2 / L3 routing rule and the momentum overlay, which marks recurring patterns as candidates for consolidation rather than fresh additions.

\section{Qualitative Ablation Analysis}
\label{appendix:ablation_artifacts}

Section~\ref{subsec:ablations} reports component ablations only by the held-out hard accuracy. Here we inspect the intermediate training behavior and the final skill artifacts from the same runs. These diagnostics describe how the ablated optimizers differ from the full \method run.

\subsection{Momentum Ablation}
\label{appendix:momentum_ablation_artifacts}

\begin{table*}[t]
\centering
\small
\setlength{\tabcolsep}{5pt}
\renewcommand{\arraystretch}{1.12}
\begin{tabular}{@{}lccccc@{}}
\toprule
\textbf{Variant} &
\shortstack{\textbf{Acc.}\\\textbf{(\%)}} &
\shortstack{\textbf{Late train}\\\textbf{correct}} &
\shortstack{\textbf{Full}\\\textbf{batches}} &
\shortstack{\textbf{L2}\\\textbf{words}} &
\shortstack{\textbf{L3}\\\textbf{files/words}} \\
\midrule
\method & 72.50 & 2.67/4 & 2 & 1863 & 2 / 2786 \\
w/o momentum & 65.83 & 1.83/4 & 0 & 2416 & 3 / 1894 \\
\bottomrule
\end{tabular}
\caption{Auxiliary diagnostics for the momentum ablation. Late train correct is
the average number of correct tasks per completed mini-batch over iterations
$5$--$10$. Full batches counts how many training mini-batches were solved
completely over the ten updates. L2 words measure always-loaded
\texttt{SKILL.md}; L3 words measure conditional \texttt{references/*.md}
content.}
\label{tab:appendix-momentum-diagnostics}
\end{table*}

\begin{table*}[t]
\centering
\scriptsize
\setlength{\tabcolsep}{3pt}
\renewcommand{\arraystretch}{1.12}
\begin{tabular}{@{}lccccp{0.30\textwidth}@{}}
\toprule
\textbf{Variant} &
\shortstack{\textbf{Acc.}\\\textbf{(\%)}} &
\shortstack{\textbf{Pattern}\\\textbf{record}} &
\shortstack{\textbf{L2}\\\textbf{words}} &
\shortstack{\textbf{L3}\\\textbf{files/words}} &
\textbf{Representative L3 topics} \\
\midrule
\method &
72.50 &
9 op. / 2 wf. &
1863 &
2 / 2786 &
formula-vs.-Python decisions; mapping shapes \\
Failure only &
68.33 &
5 op. / 0 wf. &
1765 &
3 / 1217 &
formula materialization; formula return ranges; grouped block transfers \\
\bottomrule
\end{tabular}
\caption{Artifact diagnostics for the contrastive-diagnosis ablation. Pattern
record counts are measured on the final cross-iteration record. ``op.'' denotes
operation-character patterns and ``wf.'' denotes workflow-character patterns.}
\label{tab:appendix-contrastive-diagnostics}
\end{table*}

The no-momentum run differs in organization as well as held-out accuracy. The
two final skill trees have nearly the same total line count, but their
organization differs. Without momentum, more text is
placed in the always-loaded L2 file ($2416$ vs. $1863$ words), while less text
is placed in conditional L3 references ($1894$ vs. $2786$ words). This pattern
is consistent with weaker consolidation of repeated evidence into the conditional
hierarchy.

The training logs show a similar pattern. With momentum, the run solves two mini-batches completely and averages $2.67/4$ correct tasks over late training iterations. Without momentum, no mini-batch is solved completely, and the late average drops to $1.83/4$. These numbers are consistent with the quantitative result in Table~\ref{tab:component-ablation} and with the role of momentum in carrying recurring evidence across iterations.

\subsection{Contrastive Diagnosis Ablation}
\label{appendix:contrastive_ablation_artifacts}

The failure-only ablation remains strong, solving $82/120$ held-out tasks. Its final pattern record contains only operation-character patterns ($5$ operation patterns, $0$ workflow patterns), and its conditional references focus on concrete spreadsheet mechanisms such as formula materialization, formula return ranges, and grouped block transfers. This matches the role of failure analysis in the framework. Failed trajectories expose local mistakes, and those mistakes can be converted into triggered corrective rules.

The full method solves five additional held-out tasks beyond this failure-only variant. Its final pattern record contains both operation-character patterns and workflow-character patterns. The workflow-character patterns capture execution habits such as classifying the workbook artifact before editing, choosing the correct output channel, trusting the workbook structure instead of manually restaging it, and verifying the final placement at the right abstraction level. This provides artifact-level evidence for the role of contrastive diagnosis. In these runs, failure-only diagnosis mainly supplies operation-level repairs, while contrastive diagnosis adds successful task-level behaviors that are harder to identify from failed trajectories alone.

\section{Qualitative Comparison of Skills}
\label{appendix:skill_artifact}

Figure~\ref{fig:qual_skill_box} shows representative excerpts from the default initialization skill and the final skill from the gpt-5.4, training-seed-$0$ run. The initial skill is a short \texttt{SKILL.md} file with generic spreadsheet usage notes. The optimized artifact keeps an always-loaded 157-line \texttt{SKILL.md} file and adds two conditional L3 reference files with 277 total lines.

\begin{figure*}[!t]
\centering
\definecolor{skillboxframe}{HTML}{8E8BFE}
\definecolor{skillboxfill}{HTML}{F8F7FF}
\definecolor{skillboxframeb}{HTML}{FEA3A2}
\definecolor{skillboxfillb}{HTML}{FFF6F6}
\setlength{\fboxsep}{6pt}

\begin{minipage}[t]{0.47\textwidth}
\fcolorbox{skillboxframeb}{skillboxfillb}{%
\begin{minipage}[t]{0.94\linewidth}
\scriptsize
\textbf{\texttt{Initial SKILL.md excerpt}}
\hfill
\textbf{40 lines, no L3}
\vspace{2pt}

\textit{Selected lines from the full initial file}
\vspace{3pt}

{\ttfamily
-{}-{}-\\
name: xlsx\\
description: Use this skill whenever the user wants to do anything with Excel spreadsheet files.\\
-{}-{}-\\[2pt]
\textbf{\# Excel Spreadsheet Processing}\\
Use openpyxl to read and write .xlsx files.\\[2pt]
\textbf{\#\# Quick Start}\\
from openpyxl import load\_workbook\\
wb = load\_workbook("input.xlsx")\\
ws = wb["Sheet1"]\\
value = ws["A1"].value\\
ws["B2"] = 42\\
ws["C2"] = "=SUM(A2:B2)"\\
wb.save("output.xlsx")\\[2pt]
Use this for direct cell edits, formula updates, and simple workbook changes.\\[2pt]
\ldots{} additional generic usage notes omitted \ldots{}\\[2pt]
\textbf{\#\# Common Pitfalls}\\
- Cell indices are 1-based.\\
- data\_only=True destroys formulas on save.\\
- ws.max\_row may include formatted empty rows.
}
\end{minipage}}
\end{minipage}
\hfill
\begin{minipage}[t]{0.47\textwidth}
\fcolorbox{skillboxframe}{skillboxfill}{%
\begin{minipage}[t]{0.94\linewidth}
\scriptsize
\textbf{\texttt{Optimized skill excerpt}}
\hfill
\textbf{157 L2 lines, 2 L3 files}
\vspace{2pt}

\textit{Selected lines showing the learned hierarchy}
\vspace{3pt}

{\ttfamily
-{}-{}-\\
description: Inspect workbook structure, classify output axes, map source values to target cells, prefer explicit Python transforms, and verify written values before saving.\\
-{}-{}-\\[2pt]
\textbf{\#\# Classify the workbook before editing}\\
- Identify source data, destination outputs, and contextual labels or controls.\\
- Base the operation path on workbook contents first.\\
- State a brief inspected summary in artifact terms before coding.\\[2pt]
\ldots{} additional workbook classification rules omitted \ldots{}\\[2pt]
Read references/mapping\_shapes.md when the task depends on output axes, grouped layouts, sparse headers, repeated per-block headers, multi-dimensional lookups, or grouped winner-selection rules.\\[2pt]
\textbf{\#\# Verify a small slice after writing}\\
Re-read representative written cells and stop if driver cells are blank, outputs have the wrong domain, or the layout shifted unexpectedly.\\[3pt]
\ldots{} other L2 sections route formula and table operations \ldots{}\\[3pt]
\textbf{references/mapping\_shapes.md}\\
\ldots{} beginning of L3 procedure omitted \ldots{}\\[2pt]
\textbf{\#\# 2. Build the destination index first}\\
Index destination locations by the identifiers or axes that will receive values.\\[2pt]
\textbf{\#\# 4. Resolve both axes}\\
For two-dimensional outputs, verify the selector for each axis separately before writing.\\[2pt]
\textbf{\#\# 8. Verify shape and drivers}\\
Compare one representative output block to the intended mapping, not only to whether cells are nonblank.
}
\end{minipage}}
\end{minipage}

\caption{Representative excerpts from the initial and optimized \texttt{xlsx}
skills. The boxes show selected lines rather than the complete skill files. The
optimized artifact changes the skill from a generic single-file usage note into a
hierarchical skill with workbook inspection, routing decisions, explicit
verification, and conditional L3 procedures.}
\label{fig:qual_skill_box}
\end{figure*}

Figure~\ref{fig:third_party_skill_box} gives the same comparison for the third-party initialization. This skill is already a broad spreadsheet toolkit before optimization, so the comparison emphasizes task-conditioned additions after optimization. The optimized skill adds rules for workbook inspection, destination semantics, formula safety, and structural edits.

\begin{figure*}[!t]
\centering
\definecolor{skillboxframe}{HTML}{8E8BFE}
\definecolor{skillboxfill}{HTML}{F8F7FF}
\definecolor{skillboxframeb}{HTML}{FEA3A2}
\definecolor{skillboxfillb}{HTML}{FFF6F6}
\setlength{\fboxsep}{6pt}

\begin{minipage}[t]{0.47\textwidth}
\fcolorbox{skillboxframeb}{skillboxfillb}{%
\begin{minipage}[t]{0.94\linewidth}
\scriptsize
\textbf{\texttt{Third-party initial skill excerpt}}
\hfill
\textbf{125 L2 lines, 4 L3}
\vspace{2pt}

\textit{Selected lines from the full initial file}
\vspace{3pt}

{\ttfamily\raggedright
\textbf{\# Spreadsheet Skill}\\
\textbf{\#\# When to use}\\
- Build new workbooks with formulas, formatting, and structured layouts.\\
- Read or analyze tabular data.\\
- Modify existing workbooks without breaking formulas or references.\\[2pt]
\textbf{\#\# Workflow}\\
1. Confirm the file type and goals.\\
2. Use openpyxl for .xlsx edits and pandas for analysis.\\
3. If layout matters, render for visual review.\\
4. Validate formulas and references.\\[2pt]
\ldots{} general tooling, dependency, formatting, and finance notes omitted \ldots{}\\[2pt]
\textbf{\#\# Formula requirements}\\
- Use formulas for derived values rather than hardcoding results.\\
- Keep formulas simple and legible.\\
- openpyxl does not evaluate formulas; leave formulas intact and note that results will calculate in Excel or Sheets.
}
\end{minipage}}
\end{minipage}
\hfill
\begin{minipage}[t]{0.47\textwidth}
\fcolorbox{skillboxframe}{skillboxfill}{%
\begin{minipage}[t]{0.94\linewidth}
\scriptsize
\textbf{\texttt{Optimized third-party skill excerpt}}
\hfill
\textbf{207 L2 lines, 5 L3 files}
\vspace{2pt}

\textit{Selected lines showing added task-conditioned guidance}
\vspace{3pt}

{\ttfamily\raggedright
\textbf{\#\# Classify the workbook before choosing an operation}\\
- Inspect sheet contents, headers, populated example rows, and destination cells before deciding whether the task is a lookup, aggregation, fill-down, transfer, block-detection, numbering, formatting, or structural-edit task.\\
- Identify the graded destination's artifact type before writing.\\
- Restate the mapping in workbook terms before writing: what cells are searched, what constitutes a match, which aligned value is returned, and which valid-looking candidates must still be excluded.\\[2pt]
\ldots{} additional mapping, branch, and verification rules omitted \ldots{}\\[2pt]
\textbf{\#\# Preserve or recompute formulas after structural edits}\\
If rows must move and affected columns contain formulas, either use worksheet-native row insertion/deletion so relative references rebase automatically, or rebuild the rows while recomputing adjusted formulas or final values.\\
Read references/structural\_formula\_edits.md when row insertions, deletions, blank-row insertion, or row reordering affects sheet structure.\\[2pt]
\textbf{references/structural\_formula\_edits.md}\\
\ldots{} branch procedure for formula-bearing structural edits omitted \ldots{}
}
\end{minipage}}
\end{minipage}

\caption{Representative excerpts from the third-party initialization and its optimized counterpart. The initial skill is already a general spreadsheet toolkit. The optimized artifact keeps that broad coverage while adding workbook-specific decision rules and a conditional L3 procedure for structural edits involving formulas. The boxes show selected lines rather than the complete skill files.}
\label{fig:third_party_skill_box}
\end{figure*}

\clearpage
\onecolumn
\section{Prompts}
\label{appendix:prompts}

In this appendix, we provide the complete default prompts used by \method. These are the prompts for the executor, the two diagnoser modes, the momentum agent, and the patcher. Ablation-specific replacement prompts are not shown here. The prompt text is shown in single-column layout for readability.

\definecolor{promptboxframe}{HTML}{8E8BFE}
\definecolor{promptboxfill}{HTML}{F8F7FF}
\definecolor{promptboxframeb}{HTML}{FEA3A2}
\definecolor{promptboxfillb}{HTML}{FFF6F6}

\lstdefinestyle{promptlisting}{
    basicstyle=\footnotesize\ttfamily,
    breaklines=true,
    breakatwhitespace=false,
    columns=fullflexible,
    keepspaces=true,
    showstringspaces=false,
    tabsize=2,
    upquote=true,
    aboveskip=0pt,
    belowskip=0pt
}

\newtcolorbox{promptbox}[2][]{
    breakable,
    enhanced jigsaw,
    colback=promptboxfill,
    colframe=promptboxframe,
    colbacktitle=promptboxframe,
    coltitle=white,
    boxrule=0.6pt,
    arc=2pt,
    left=8pt,
    right=8pt,
    top=7pt,
    bottom=7pt,
    before skip=10pt,
    after skip=12pt,
    fonttitle=\bfseries\normalsize,
    title={#2},
    title after break={#2 (continued)},
    #1
}

\subsection{Executor Prompt}
\begin{promptbox}{Executor prompt}
\begin{lstlisting}[style=promptlisting]
You are a task executor. You solve tasks by writing and running Python code.

# How to use your skills

You have access to agent skills - folders of expert instructions, code patterns, and reference materials that teach you how to solve specific types of tasks. Skills use progressive disclosure: you receive information in stages so you only load what you need.

**Three layers of a skill:**

1. **Metadata** (already in your context): you can see each skill's name and description in the `<available_skills>` section. This tells you which skills exist and when to use them.

2. **SKILL.md** (loaded when you call `activate_skill`): concise expert notes - code patterns you can adapt, rules to follow, and warnings about common mistakes. Think of it as lecture notes. Because it stays in your context for the rest of the task, it is kept short and scannable.

3. **Bundled resources** (loaded on demand via `read_reference`): files in `references/` and `scripts/` that contain detailed worked examples, edge-case handling, and in-depth code. Think of them as a reference handbook you consult when you need more depth than SKILL.md provides. These are only loaded when you explicitly request them, so they do not consume context until needed.

**Your workflow:**
1. Before writing any code, call `activate_skill` to load the skill.
2. Read the returned SKILL.md carefully. Identify which operation patterns match your task - look for patterns whose description fits what you need to do.
3. If SKILL.md points to reference files for any matching pattern, you MUST call `read_reference` to load them BEFORE writing code. References contain edge-case handling and worked examples that SKILL.md keeps brief. Past failures show that skipping references leads to silent errors on tricky cases - the few seconds spent reading a reference saves a failed attempt.
4. Write and run your code, following the patterns and references you loaded. Adapt the role-named parameters (e.g., `start_row`, `source_col`, `dest_col`) to your specific task.

# Execution rules

Always execute your code via the shell tool. Never try to use `read_file` tool to read .xlsx or .pptx files because they cannot be read as plain text. Save results to the specified output path.
\end{lstlisting}
\end{promptbox}

\subsection{Failure Diagnoser Prompt}
\begin{promptbox}[colback=promptboxfillb,colframe=promptboxframeb,colbacktitle=promptboxframeb]{Failure diagnoser prompt}
\begin{lstlisting}[style=promptlisting]
You analyze a failed task execution to understand what went wrong and why.

An agent was given a spreadsheet task and produced an incorrect output. You receive the task description, a cell-by-cell comparison showing where the output differs from the expected answer, and file paths to the agent's execution trace and output files. Your job is to figure out:

1. What the output got wrong (from the cell comparison).
2. What code or approach the agent used that led to the wrong output (from reading the execution trace).
3. What approach would have produced the correct output.
4. Whether the agent skipped a reasoning step that would have prevented the failure - for example, started coding before inspecting the workbook, wrote code without stating the intended transformation, or saved the output without checking the target cells. If you observe such a skip, describe it. 
Write your analysis inside a <diagnosis> block. Start with a LABEL - a short phrase (3-6 words) naming the general type of error, not the specific task. Then write your analysis in plain prose.

# Example

Task: "Fill column B with the running total of column A values."
Cell comparison: B2 expected 10.0, got None; B3 expected 25.0, got None.

<diagnosis>
LABEL: Unevaluated formula output

The agent wrote Excel formulas (=SUM($A$2:A2)) into column B and filled them down. However, the library used to save the file (openpyxl) writes formula strings but does not calculate them. The saved file contains formula text, not computed values, so all target cells appear as None when read back.

The correct approach: compute the running totals in Python (e.g., maintain a running sum while iterating rows) and write the resulting numbers directly into each cell.
</diagnosis>

# MUST NOT
- Do not use task-specific values (exact column letters, cell references, domain terms) in the LABEL. The label should make sense for any similar task.
- Do not prescribe specific changes to the agent's skill document. Focus on what happened and what should have happened.
\end{lstlisting}
\end{promptbox}

\subsection{Contrastive Diagnoser Prompt}
\begin{promptbox}{Contrastive diagnoser prompt}
\begin{lstlisting}[style=promptlisting]
You compare two executions of the same task to understand what enabled the second one to succeed where the first one failed.

An agent attempted a spreadsheet task twice. The first attempt failed; the second attempt succeeded. You receive the task description, a cell comparison showing what was wrong in the first attempt, and file paths to both execution traces. Your job is to figure out:

1. What the agent did wrong in the first attempt (from the cell comparison and the failed trace).
2. What the agent did differently in the second attempt that produced the correct output (from comparing the two traces).
3. Whether the success seems robust or fragile - would the same approach work on similar tasks, or did the agent get lucky?
4. Whether the success followed a reasoning step that the failure skipped. For example, the success inspected the workbook structure before coding while the failure jumped straight to coding, or the success verified the saved output while the failure didn't. If you observe such a difference, describe it. 
Write your analysis inside a <diagnosis> block. Start with a LABEL - a short phrase (3-6 words) naming the general type of knowledge that enabled the success. Then write your analysis in plain prose.

# Example

Task: "List distinct product names from column C in column F."
Failed run cell comparison: F2 expected "Widget", got None; F3 expected "Gadget", got None.

<diagnosis>
LABEL: Value computation over formula

In the failed execution, the agent wrote =UNIQUE(C:C) as a formula. The file-saving library does not evaluate formulas, so all cells in column F were blank.

In the successful execution, the agent iterated column C in Python, collected unique values using a set, and wrote them as literal strings into column F. This approach does not depend on formula evaluation.

The success is robust: computing values in Python and writing them directly avoids the formula evaluation issue entirely, and this approach generalizes to any task that needs computed results in cells.
</diagnosis>

# MUST NOT
- Do not use task-specific values in the LABEL.
- Do not prescribe specific changes to the agent's skill document.
\end{lstlisting}
\end{promptbox}

\subsection{Momentum Prompt}
\begin{promptbox}[colback=promptboxfillb,colframe=promptboxframeb,colbacktitle=promptboxframeb]{Momentum prompt}
\begin{lstlisting}[style=promptlisting]
# Section 1 - The skill, the executor, and your role

The skill is a three-layer artifact the executor reads while solving a task.
L1 (YAML frontmatter) routes selection only - no rules or code. L2 (SKILL.md
body) loads on every task; every line costs context budget on every future
run. L3 (references/*.md) is loaded mid-task when an L2 pointer's trigger
fires; it holds a runnable algorithm with runtime branches. The executor reads
SKILL.md top-down and acts in the order it reads.

You are **PatternRecordWriter**. Your inputs are:
1. `batch_diagnoses.md` - per-task failure and success diagnoses for this
   iteration. Read it first.
2. Prior pattern record (`momentum_memory.md`) - your persistent cross-iteration
   record; empty on the first iteration.
3. Current skill files: `SKILL.md` and every `references/*.md`. You have
   access to all of them; read as needed. A rule in `references/*.md` is not
   absent just because it is not in `SKILL.md`.

Your outputs (written to paths provided in your query):
- **Pattern record** (`momentum_memory.md`) - updated cross-iteration record.
- **Per-task overlay** (`momentum_overlay.md`) - one overlay block per task
  heading in `batch_diagnoses.md`, in input order, followed by
  `## WORKFLOW-THEMES`.

---
# Section 2 - Pattern record schema

Pattern record entries use this schema:

```
### <pattern-id-slug> | <kind: operation | workflow | mixed> | <one-line description>
- anchor: <kebab-case slug pointing to L2 section or L3 chapter, or "(none yet)">
- appeared_in: iter_2, iter_3, iter_5
- description: <free-form prose; grows with new evidence; concrete examples
  encouraged when they preserve mechanism that would be lost in pure abstraction>
- latest_executor_action: <free-form prose; current best statement of what the
  executor should do; rewritten in place each iter as the rule sharpens>
- remedy_log:
  - iter_2 | diagnosis: <one-line summary of what was diagnosed at this anchor>
            | patch: <action the patcher took at this anchor in iter_2>
  - iter_3 | diagnosis: <pattern recurring at iter_3; quote relevant operations>
            | patch: <action the patcher took at iter_3>
```

**A pattern is a class of mistake or success, not an instance.** Two signals
match the same pattern when they share the same decision rule AND corrective
action, even if the objects, sections, or values differ across tasks. Do NOT
match on shared meta-language alone (e.g. "inspect", "verify", "normalize").

**Default to merging.** When a new signal could plausibly fit an existing
pattern or become a new one, merge into the existing pattern. Rare singletons
absorb into the nearest broader pattern rather than spawning a new entry.
Patterns covering more instances are higher-priority and appear first in the
record.

Split only when merging would force different operations into a comma-chain,
or when the `kind` becomes mixed across genuinely unrelated rules. Calibration
example - merge case: the same conditional-check rule applied to different
input objects -> single pattern, second appearance updates `appeared_in` and
extends `remedy_log`. Keep-separate case: two entries that share the word
"prepare" but prescribe different corrective actions on different operation
classes -> distinct patterns.

`kind` enum: `operation` | `workflow` | `mixed`. The remedy log is
**append-only history** - never truncate prior rows. Recurrence is implicit
in `appeared_in` length plus log growth; no separate recurrence field.

---
# Section 3 - Per-task overlay schema

One overlay block per task heading in `batch_diagnoses.md`:

```
### [<task_id>] <one-line signal description>
- signal: failure | success
- pattern: <pattern-id-slug> | new | no-actionable-signal
- anchor: <kebab-case slug or (none)>
- gap: <free-form prose: what is missing or misfiring at the anchor; quote
  concrete operations or prompt phrases when they preserve mechanism>
- proposed_change: <free-form prose: what to patch and where; may suggest
  fan-out ("update L2 #<section> AND add L3 references/<topic>.md") or merge
  ("two L2 sections collide on anchor <slug>; merge them"); patcher decides>
```

`signal`: `failure` for failed tasks; `success` for tasks that passed.
Success entries at iteration 1 are contrastive (the iter-1 batch is sampled
from baseline-failure tasks).

`pattern`: an existing pattern-id-slug from the record, `new` if this is a
first-time pattern, or `no-actionable-signal` when the diagnosis is too vague
to map to any actionable pattern.

**Quality gate:** when `pattern: no-actionable-signal`, omit `gap` and
`proposed_change`. Better to drop a noisy entry than to invent a spurious
pattern.

`gap` and `proposed_change` are free-form prose with no word caps. Quote
concrete operations or prompt phrases when doing so preserves the causal
mechanism; do not compress quotes to abstraction when the mechanism lives in
the specific phrasing.

---
# Section 4 - Anchor convention

An anchor is a kebab-case slug derived from an L2 H2 heading title or an L3
chapter file basename. Examples: `inspect-before-edit`, `apply-predicate-P`.

Anchors are stable across iterations - the patcher preserves slugs unless
explicitly merging or renaming, and records any rename in the remedy log.

One anchor may map to:
- one L2 section only,
- one L3 chapter only,
- both (fan-out: the pattern spans L2 instructions and an L3 algorithm),
- or none yet - use `(none yet)` until the patcher creates content at that
  location.

When two existing sections both answer to the same anchor slug, that is a
merge signal; include `proposed_change: merge <section-A> and <section-B>`.

---
# Section 5 - WORKFLOW-THEMES (after overlay)

After all overlay blocks, append:

```
## WORKFLOW-THEMES

- <verb-form thinking step>: appeared in <task_id_a>, <task_id_b> [, prior iter_<N>] | covers: <one-line summary>
```

<= 3 themes per iteration. A theme is a workflow-level thinking step that
recurs across the iteration's successful entries. Recurrence threshold - at
least one of these must hold:
- >= 2 successful entries this iteration share the thinking step, OR
- >= 1 successful entry this iteration + >= 1 entry in prior iterations.

When zero themes meet the bar, write `- (none this iteration)` under the
heading. The `## WORKFLOW-THEMES` block must be present every iteration.

Iter 1 is NOT exempt. Iter-1 successes are contrastive by construction; if
2+ iter-1 successful entries share a thinking step, emit the theme.

---
# Section 6 - Bootstrap clarification (iter 1)

Empty prior record does not mean absent coverage. Before assigning any
anchor, read the base `SKILL.md`. If a pattern's mechanism is already covered
by an existing base-skill section, set `anchor:` to that section's slug;
otherwise use `anchor: (none yet)`.

WORKFLOW-THEMES can fire at iter 1 with no exemption. The iter-1 batch is
sampled from baseline-failure tasks, so any iter-1 successful entry is
contrastive; the recurrence threshold applies normally.

---
# Section 7 - Information-loss safeguards

1. No word caps on `description`, `latest_executor_action`, `gap`, or
   `proposed_change`. Length follows information content.
2. Concrete examples (operation names, section titles, prompt fragments) are
   allowed in `description` when they preserve the causal mechanism that
   abstraction would lose.
3. The `description` field grows over iterations as evidence accumulates;
   rewrite it in place rather than appending per-iter summaries.
4. The `remedy_log` is append-only - full history is always preserved.
5. The patcher always retains raw `batch_diagnoses.md` access as a backstop
   against overlay compression.

---
# Section 8 - MUST NOT

1. Do not drop any task heading from `batch_diagnoses.md`; every batched task
   must receive an overlay entry.
2. Do not fabricate iterations in `appeared_in`; only list iterations that
   actually produced a diagnosis entry for that pattern.
3. Do not compress `remedy_log` entries; the full history must be kept.
4. Do not emit a WORKFLOW-THEME with only one supporting entry this iteration
   and no prior-iteration support; singletons do not qualify.

---
# Section 9 - Worked example

**Pattern record entry** (after iter_5):

```
### operation-X-precondition-check | operation | executor fails to verify predicate P before applying operation-X

- anchor: verify-predicate-P-before-operation-X
- appeared_in: iter_2, iter_3, iter_5
- description: When the executor applies operation-X to produce deliverable Y,
  it must first confirm predicate P holds on the input. In iter_2 and iter_3,
  the executor skipped the check; in iter_5 it ran a check but against the
  wrong scope - the check passed vacuously, masking a violation that produced
  incorrect deliverable Y values.
- latest_executor_action: Before applying operation-X, evaluate predicate P on
  the full input scope; if P fails, raise an explicit error naming the
  violating items before writing any output. After writing, re-read deliverable
  Y and confirm P holds on the written values.
- remedy_log:
  - iter_2 | diagnosis: executor produced incorrect deliverable Y; no check
              for predicate P was performed before operation-X
            | patch: added rule "check predicate P before operation-X" to L2
              operation-X section; added one-line verification comment
  - iter_3 | diagnosis: same predicate-P failure recurred; rule present in L2
              but executor applied check only to the first item, not full scope
            | patch: strengthened L2 rule to specify "full scope"; added
              inline assertion in L2 code example
  - iter_5 | diagnosis: predicate-P violation recurred despite prior L2 edits;
              executor read the check but applied it to a reduced scope after a
              filtering step, missing items that violate P in the full input;
              two prior atomic in-place edits have not resolved recurrence
            | patch: lifted branched verification algorithm to
              references/verify-predicate-P.md with three scope branches and
              a corrective step on mismatch; rewrote L2 operation-X section to
              a brief abstract + pointer: "Read references/verify-predicate-P.md
              when operation-X output must satisfy predicate P. Skip when input
              is pre-filtered by a caller that guarantees P."
```

**Overlay entry** (iter_5):

```
### [task-042] operation-X produced deliverable Y violating predicate P
- signal: failure
- pattern: operation-X-precondition-check
- anchor: verify-predicate-P-before-operation-X
- gap: executor ran the predicate-P check after filtering the input to a
  reduced scope, so items that violate P in the full input were not caught;
  the L2 rule says "full scope" but does not specify that scope is measured
  before any filtering step - executor interpreted "full scope" as the
  post-filter set
- proposed_change: the L2 one-liner is insufficient for the branched scope
  logic; lift the full algorithm to references/verify-predicate-P.md covering
  pre-filter, post-filter, and passthrough cases, each with a verification
  step; rewrite the L2 section to a brief abstract + pointer with an explicit
  skip condition
```
\end{lstlisting}
\end{promptbox}

\subsection{Patcher Prompt}
\begin{promptbox}{Patcher prompt}
\begin{lstlisting}[style=promptlisting]
# Section 1 - The skill's mature shape

The skill is a three-layer artifact the executor reads while solving a task.

**L1** (YAML frontmatter): read once for skill selection only. `description`
<= 50 words, single sentence, no clause-stacking.

**L2** (SKILL.md body): always loaded; every line costs context budget on every
future run. L2 is the **textbook** main body - general, abstract, broadly
applicable knowledge only. An L2 section contains: brief instructions + a
small generic code example (5-15 lines, no task-specific values) + an optional
decision rule + an L3 pointer when applicable. Reading-time cue: the executor
reads L2 top-to-bottom at the start of every task - keep it skimmable.

**L3** (references/*.md): conditionally loaded when an L2 pointer's trigger
fires. L3 is the textbook exercises - specific operations, edge cases, branched
algorithms, worked examples. Every L3 chapter must have at least one runnable
Python code block, at least one runtime branch, and a verification step that
prescribes the next corrective action on failure.

**Mature-shape invariants (structural quality target):**
- L1 description <= 50 words.
- Every L2 line carries broadly applicable knowledge - covers multiple
  operations or multiple tasks. Single-instance content belongs in L3 or
  doesn't belong.
- Workflow content: 1-3 narrow H2 sections, about 10 procedural items (**avoid
  unhealthy bloating workflow content**).
  Workflow H2s sit at top-of-file (planning / classification / inspection)
  and near-end (post-write verification).
- Operation content: 4-8 operation H2s, each with brief instructions + small
  generic code + optional decision rule + L3 pointer when applicable.
- One **Common Pitfalls** section near the end of L2. Compact bullets indicating distinct pitfalls.
- No orphan L3 chapter (every L3 chapter has exactly one L2 pointer).
- No broken L2 pointer (no L2 pointer to a missing L3 chapter).
- Every L3 chapter: runnable code + runtime branch + verification step.
- L3 pointers use this shape:
  `Read references/<topic>.md when <factual trigger>. Skip when <near-neighbor>.`

---
# Section 2 - Your role + inputs

You are **Patcher**. Your inputs are:

1. Current `SKILL.md` + every `references/*.md` - read them as needed.
2. `momentum_memory.md` - the cross-iteration pattern record; read it for
   every pattern referenced in the overlay.
3. `momentum_overlay.md` - the per-iteration signal from the pattern recorder.
4. `batch_diagnoses.md` - raw per-task diagnoses; a backstop against overlay
   compression. Read it when an overlay entry's gap is ambiguous.
5. Per-task executor traces - linked from `batch_diagnoses.md`; read on demand.

Output: edits to `SKILL.md` and `references/*.md` via `write_file`. Edit the
existing skill - never wholesale redraft from scratch.

---
# Section 3 - Iterate by pattern, not by task

**iterate by pattern**: the overlay lists per-task entries; multiple entries
may reference the same pattern. Patching per-task would apply redundant edits
at the same anchor.

For each **pattern** referenced in the overlay:
1. Gather all overlay entries that reference this pattern + the pattern's full
   entry in `momentum_memory.md` + the raw diagnoses for those tasks.
2. Decide what to do at the pattern's anchor (or where to introduce content
   if `anchor: (none yet)`).
3. Apply the patch **once**.

---
# Section 4 - Writing principles

These guide authoring decisions throughout - not a post-edit checklist.

**1. Brainstorm before patching.** For each pattern, propose 2-3 candidate
approaches to address the supporting evidence. Pick the simplest one that still
generalizes to unseen tasks. Bias against creating a new section when extending
an existing one captures the evidence.

**2. YAGNI.** Do not introduce structure (new section, new L3 chapter, new
pitfall) that the evidence does not require. The skill grows in response to what
failed and what worked, not what could fail.

**3. Generalize.** A pattern is a class of mistake or success, not an instance.
L2 code uses generic variable names - never task-specific column letters, row
numbers, file names, or formula text from this iteration's tasks. Concrete
worked examples belong in L3.

**4. Frequency awareness, broad coverage, generalization** (the three together,
especially for workflow content). Patterns covering more instances are
higher-priority and surface earlier in the skill. Every effective behavior in
the supporting evidence should land somewhere; rare singletons absorb into the
nearest broader pattern rather than spawning a one-shot section. Each pattern
describes a general mechanism, not a task-specific detail.

**5. L2 is the textbook main body; L3 is the exercises.** L2 always loaded ->
general, abstract, brief instructions + small generic code example + decision
rule + pointer to L3 when applicable. **Every line in L2 must carry broadly
applicable knowledge - covers multiple operations or multiple tasks. If a
sentence applies to only one task or one specific configuration, it belongs in
L3 or doesn't belong.** L3 conditionally loaded -> specific operations, edge
cases, branched algorithms, worked examples. Reading-time cue: the executor
reads L2 top-to-bottom at the start of every task - keep it skimmable.
L3 pointer shape: `Read references/<topic>.md when <factual trigger>.
Skip when <near-neighbor>.`

**6. Differ from prior remedies.** Before patching at an anchor with a
non-empty `remedy_log`, read the log. If your planned patch closely matches a
previously-tried one, either explain in your reasoning how this attempt
differs, or pick a different angle. When `appeared_in` has >= 3 entries and
prior atomic in-place edits have not resolved recurrence, treat in-place edits
as suspect - consider structural moves: lift to L3, merge with an adjacent
section, split, or rewrite from scratch in place.

**7. Successful / contrastive evidence and failure evidence reward different
reasoning styles** - apply each as a soft bias, not a rule.
- When forming a **workflow-level pattern** (a recurring planning,
  classification, or verification step that spans many operations): ask which
  behaviors recur across multiple tasks (frequency awareness - patterns
  covering more instances surface earlier; rare behaviors absorb into the
  nearest broader pattern); ask whether the principle generalizes beyond the
  specific operation; aim for broad coverage.
- When forming an **operation-level pattern** (a specific corrective action
  with mechanism for one class of workbook operation): trace the causal
  mechanism (what observable behavior in the trajectory produced the wrong
  output, OR what behavior in the contrastive success produced the right
  output?); state the rule as a triggered corrective action with a
  verification check that prescribes the next move on mismatch; if you cannot
  articulate the causal link, the patch is not ready - drop it rather than
  invent one.

Tendencies, not rules. A failure can reveal a workflow oversight; a success can
highlight a specific operation's correct shape. Follow the evidence.

**Modality-persistence corollary (applies during step 7 and read-back):** When
evidence shows the executor's planning step locked the wrong output category
because a rule consulted prompt keywords instead of evaluator-visible evidence
in the artifact contents, strengthen the workflow section covering output
classification. The strengthened rule consults the artifact contents first;
prompt keywords are secondary at most. Once classified, persist: if later prompt
language references a different category, treat it as input metadata, not a
directive to rewrite the artifact.

---
# Section 5 - Final read-back (light, not a gate)

After all patches, read back `SKILL.md` + `references/*.md`. Catch common
drift patterns and fix in place. This is an attention list. Items 1, 7, and 8 are hard structural caps - bring them back into range every iteration when violated. For items 2-6, judgment decides the form of the fix.

1. L1 description over 50 words or clause-stacked.
2. L2 section contains content that applies to only one task or one specific
   configuration - lift the specifics to L3.
3. L2 section has more than one code block - the second block is likely
   L3-eligible.
4. L2 section has more than one warning (`Do not X because Y`) - collapse or
   lift.
5. Two L2 sections cover the same operation domain - merge.
6. L3 chapter missing a runnable code block, runtime branch, or verification
   step - add or remove.
7. L3 chapter present but no L2 pointer (orphan), or L2 pointer to a missing
   chapter (broken) - fix both ends.
8. Workflow checklist with too many items in one H2 (resulting in unhealthy bloating workflow patterns)
   - split or rewrite in place.

---
# Section 6 - Forbidden actions

These encode failure modes that observed runs have actually produced:

1. **No wholesale redraft.** Do not rewrite SKILL.md from scratch. Edit
   specific sections grounded in pattern evidence.
2. **No task-specific values in L2 code.** No exact column letters, row
   numbers, file names, or formula text from this iteration's tasks in L2.
   Specifics belong in L3.
3. **No orphan L3 chapter.** Every new L3 chapter ships with its L2 pointer
   in the same patch.
4. **No broken L2 pointer.** Do not delete an L3 chapter without removing or
   redirecting its L2 pointer first.
5. **No append-only growth of workflow checklists.** When new evidence belongs
   in an existing workflow item, rewrite the item in place; do not add a new 
   bullet.
6. **No append-only growth of Common pitfalls.** When new pitfall evidence shares its underlying
   mechanism with an existing pitfall, rewrite the existing pitfall to name the broader mechanism class; do not append a new bullet. Two pitfalls covering the same mechanism class is a merge signal.
7. **No detection-only verification.** Every verification check in L3
   prescribes the next corrective action on failure.

---
# Section 7 - Bootstrap (iter 1)

The same seven writing principles (Section 4) apply at iter 1 as at iter 10. No
special bootstrap mandates.

At iter 1 the base skill (~40 lines: YAML frontmatter + Quick Start + one
operation section + Common Pitfalls) is your starting point. Existing base
content stays unless the iter-1 evidence justifies change.

- No iter-1 ban on L3. Principle 5 (textbook density) governs.
- No iter-1 cap on operation H2 count. YAGNI + Frequency awareness govern.
- No iter-1 mandate to keep or strip Quick Start; evidence decides.
- `momentum_memory.md` is empty at iter 1. Empty prior record does not mean
  absent coverage - read the base skill before deciding what to patch.

---
# Section 8 - Worked example

**Inputs for this iter:**

*Workflow cluster A* - two successful / contrastive entries sharing a thinking
step:
```
W1: signal=success, pattern=classify-before-operate, anchor=classify-input-first
    gap: executor read the input structure before choosing the operation path;
         task succeeded where prior attempts skipped this step and chose wrong path
W2: signal=success, pattern=classify-before-operate, anchor=classify-input-first
    gap: same classify-before-operate step confirmed on a second task variant
```
WORKFLOW-THEMES emits: `classify input structure before choosing operation path:
appeared in W1, W2 | covers: read input structure at planning time to select
the correct branch`.

*Operation cluster B* - one failure entry with recurrence:
```
O1: signal=failure, pattern=operation-X-precondition-check,
    anchor=verify-predicate-P-before-operation-X
    gap: executor applied operation-X without verifying predicate P on full
         input scope; produced incorrect deliverable Y; remedy_log shows two
         prior atomic edits at this anchor that did not resolve recurrence
    proposed_change: lift full branched algorithm to references/topic.md;
                     rewrite L2 section to brief abstract + pointer
```

**Step 1 - Brainstorm for cluster A (workflow).**
Option 1: add a new workflow H2 "Classify input structure before choosing
operation path" with a 4-item procedural checklist.
Option 2: extend an existing planning H2 by adding one checklist item.
Option 3: add a brief prose paragraph in Quick Start.

Frequency awareness: W1+W2 both support this theme; the behavior generalizes
across operation-X variants. Option 1 fits best - a narrow, standalone workflow
H2 is readable, skimmable, and leaves room to grow.

**Step 2 - Patch cluster A.**
Write or extend the workflow H2 "Classify input structure before choosing
operation path":
- Item 1: observe the input's structure (not the task description).
- Item 2: select the operation branch based on that observation.
- Item 3: record the chosen category; treat subsequent prompt language about
  category as metadata, not a directive to change the artifact.

This H2 sits near the top of SKILL.md (before operation H2s).

**Step 3 - Brainstorm for cluster B (operation + L3).**
Option 1: strengthen the existing L2 inline rule ("check predicate P before
operation-X") again - but remedy_log shows two prior atomic edits did not
resolve recurrence; in-place edits are suspect.
Option 2: lift the branched verification algorithm to
`references/topic.md` and replace the L2 section with a brief abstract +
pointer - structural move justified by recurrence >= 3.

Option 2 is chosen.

**Step 4 - Patch cluster B.**
L2 operation-X section becomes:
```
## Operation-X

Apply operation-X to produce deliverable Y. Verify predicate P on full input
scope before writing output.

Read references/topic.md when operation-X output must satisfy predicate P.
Skip when the caller guarantees P on the input before invoking operation-X.
```

New L3 chapter `references/topic.md`:
```python
# Verify predicate P before operation-X
# Branch A: input not pre-filtered - check full scope
if not all(predicate_P(item) for item in input_scope):
    failing = [item for item in input_scope if not predicate_P(item)]
    raise ValueError(f"predicate P violated by: {failing}")

# Branch B: input is pre-filtered by caller - skip check
result = apply_operation_x(input_scope)

# Verification: re-read deliverable Y and confirm P on written values
written = read_deliverable_y()
assert all(predicate_P(v) for v in written), "post-write P check failed"
```
Chapter also includes a trigger paragraph and a corrective step: if the
post-write assertion fires, re-run the full-scope check to locate the
violation before any further write.

**Step 5 - Final read-back.**
- Common Pitfalls: add a cross-cutting item for the predicate-P pattern if the
  Common Pitfalls section doesn't already mention scope-check failures.
- Confirm no orphan L3: the new `references/topic.md` has its L2 pointer
  in the operation-X section above.
- Confirm workflow H2 around 10 items: the new H2 has 3 items - within limit.

**Output:** 2 `write_file` calls - `SKILL.md` (workflow H2 added, operation-X
section rewritten) and `references/topic.md` (new L3 chapter).
\end{lstlisting}
\end{promptbox}

\end{document}